\documentclass[10pt,twocolumn,letterpaper]{article}

\usepackage{cvpr}
\usepackage{times}
\usepackage{epsfig}
\usepackage{graphicx}
\usepackage{amsmath}
\usepackage{amssymb}
\usepackage[latin1]{inputenc}
\usepackage[francais,english]{babel}    
\usepackage{amssymb}
\usepackage{amsmath}
\usepackage{amsthm}
\usepackage{bbm}
\usepackage{graphicx}
\usepackage{color}
\usepackage[T1]{fontenc}
\usepackage{yfonts}
\usepackage{placeins}
\usepackage{array}
\usepackage{stmaryrd}
\usepackage{wasysym}
\usepackage{stackrel}
\usepackage{multirow}

\newcommand{\beq}{\begin{eqnarray}}
\newcommand{\eeq}{\end{eqnarray}}
\newcommand{\bit}{\begin{itemize}}
\newcommand{\eit}{\end{itemize}}
\newcommand{\ben}{\begin{enumerate}}
\newcommand{\een}{\end{enymerate}}

\newcommand{\bfg}{
\begin{figure}[h]
\begin{center}}
\newcommand{\efg}{
\end{center}
\end{figure}
\FloatBarrier}

\def\vect#1{\mbox{\boldmath $#1$}}
\def\x{{\mathbf {x}}}

\def\1{{\mathbf 1}}

\def\X{{\mathbf X}}

\def\alphab{{\boldsymbol\alpha}}

\def\y{{\vect{y}}}

\def\D{{\mathbf D}}

\def\Real{{\mathbb R}}

\def\A{{\mathbf A}}

\def\st{~~\text{s.t.}~~}
\def\defin{\stackrel{\vartriangle}{=}}

\newcommand{\R}{\mathbb{R}}

\newcommand{\bfx}{\mathbf {x}}
\newcommand{\bfX}{\mathbf{X}}

\newcommand{\bfE}{\mathbf{E}}
\newcommand{\bfD}{\mathbf{D}}
\newcommand{\bfDD}{\boldsymbol{\Delta}}
\newcommand{\bfd}{\mathbf{d}}
\newcommand{\bfa}{\boldsymbol{\alpha}}

\newcommand{\bfA}{\mathbf{A}}

\newcommand{\bfR}{\mathbf{R}}
\newcommand{\bfG}{\boldsymbol{\Gamma}}

\newcommand{\argmin}{\operatornamewithlimits{argmin}}

\renewcommand{\Im}{\operatorname{Im}}

\newcommand{\diag}{\operatorname{diag}}

\newcommand{\Id}{\operatorname{Id}}

\renewcommand{\geq}{\geqslant}
\renewcommand{\leq}{\leqslant}

\usepackage[pagebackref=false,breaklinks=true,letterpaper=true,colorlinks,bookmarks=false]{hyperref}

\cvprfinalcopy 

\def\cvprPaperID{1464} 

\begin{document}

\title{Sparse Image Representation with Epitomes}

\author{Louise Beno\^it$^{1,3}$
\\
\and
Julien Mairal$^{2,3}$\\
\and
Francis Bach$^{2,4}$\\
\and
Jean Ponce$^{1,3}$\\
\and
\\
$^1$École Normale Supérieure\\
$45$, rue d'Ulm,\\
$75005$ Paris, France.\\
\and
\\
$^2$INRIA\\
$23$, avenue d'Italie,\\
$75013$ Paris, France.
}

\maketitle

\footnotetext[3]{WILLOW project-team, Laboratoire d'Informatique de l'École Normale Sup\'erieure, ENS/INRIA/CNRS UMR $8548$.}
\footnotetext[4]{SIERRA team, Laboratoire d'Informatique de l'École Normale Sup\'erieure, ENS/INRIA/CNRS UMR $8548$.}

\begin{abstract}
Sparse coding, which is the decomposition of a vector using only a few basis elements, is widely used in machine learning and image processing. The basis set, also called dictionary, is learned to adapt to specific data. This approach has proven to be very effective in many image processing tasks. Traditionally, the dictionary is an unstructured ``flat'' set of atoms.  In this paper, we study structured dictionaries~\cite{aharon2} which are obtained from an epitome~\cite{jojic}, or a set of epitomes. The epitome is itself a small image, and the atoms are all the patches of a chosen size inside this image. This considerably reduces the number of parameters to learn and provides sparse image decompositions with shift-invariance properties.  We propose a new formulation and an algorithm for learning the structured dictionaries associated with epitomes, and illustrate their use in image denoising tasks.
\end{abstract}

\section{Introduction}

Jojic, Frey and Kannan~\cite{jojic} introduced in $2003$ a probabilistic generative image model called an {\em epitome}. Intuitively, the epitome is a small image that summarizes the content of a larger one, in the sense that for any patch from the large image there should be a similar one in the epitome. This is an intriguing notion, which has been applied to image reconstruction tasks~\cite{jojic}, and epitomes have also been extended to the video domain~\cite{cheung}, where they have been used in denoising, superresolution, object removal and video interpolation. Other successful applications of epitomes include location recognition~\cite{ni} or face recognition~\cite{chu}.

\begin{figure}
\centerline{
\includegraphics[width=\linewidth]{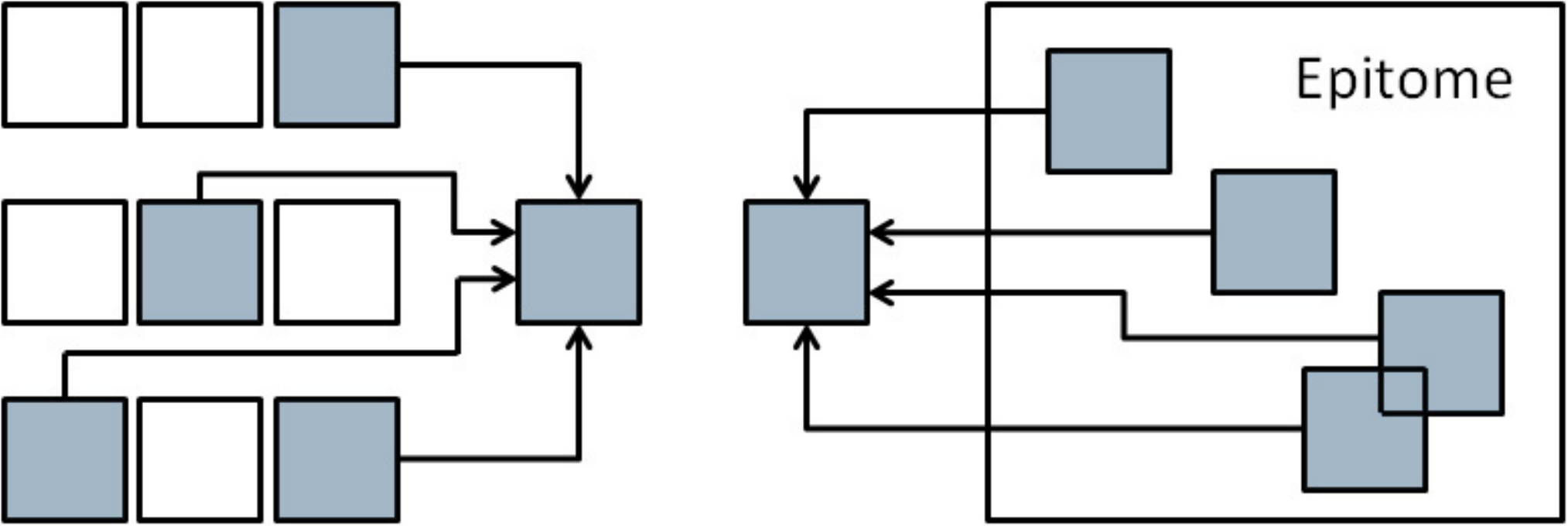}}
\caption{A ``flat'' dictionary (left) vs. an epitome (right). Sparse coding with an epitome is similar to sparse coding with a flat dictionary, except that the atoms are extracted from the epitome and may overlap instead of being chosen from an unstructured set of patches and assumed to be independent one from each other.}
\label{fig:epitome}
\vspace{-.5cm}
\end{figure}

Aharon and Elad~\cite{aharon2} have introduced an alternative formulation within the sparse coding framework called \emph{image-signature dictionary}, and applied it to image denoising. Their formulation unifies the concept of epitome and dictionary learning~\cite{elad,field} by allowing an image patch to be represented as a sparse linear combination of several patches extracted from the epitome (Figure \ref{fig:epitome}). The resulting sparse representations are highly redundant (there are as many dictionary elements as overlapping patches in the epitome), with dictionaries represented by a reasonably small number of parameters (the number of pixels in the epitome). Such a representation has also proven to be useful for texture synthesis~\cite{peyre}.

In a different line of work, some research has been focusing on  learning shift-invariant dictionaries~\cite{mailhe,thiagarajan}, in the sense that it is possible to use dictionary elements with different shifts to represent signals, exhibiting patterns that may appear several times at different positions.  While this is different from the image-signature dictionaries of Aharon and Elad~\cite{aharon2}, the two ideas are related, and as shown in this paper, such a shift invariance can be achieved by using a collection of smaller epitomes.  In fact, one of our main contributions is to unify the frameworks of epitome and dictionary learning, and establish the continuity between dictionaries, dictionaries with shift invariance, and epitomes.

We propose a formulation based on the concept of epitomes/image-signature-dictionaries introduced by \cite{aharon2,jojic}, which allows to learn a collection of epitomes, and which is generic enough to be used with epitomes that may have different shapes, or with different dictionary parameterizations. 
We present this formulation for the specific case of image patches for simplicity, but it applies to spatio-temporal blocks in a straightforward manner. 

The following notation is used throughout the paper: 
we define for $q \geq 1$ the \mbox{$\ell_q$-norm} of a vector~$\x$ in~$\R^m$ as $\|\x\|_q \defin (\sum_{j=1}^m |x_j|^q)^{{1}/{q}}$, where~$x_j$ denotes the $j$-th coordinate of~$\x$.  if $\X$ is a matrix in $\Real^{m \times n}$, $\x^i$ will denote its $i^{th}$ row, while $\x_j$ will denote its $j^{th}$ column. As usual, $x_{i,j}$ will denote the entry of $\X$ at the $i^{th}$-row and $j^{th}$-column. We consider the Frobenius norm of~$\X$: $\|\X\|_F \defin (\sum_{i=1}^m \sum_{j=1}^n x_{i,j}^2)^{{1}/{2}}$.

This paper is organized as follows: Section \ref{sec:approach} introduces our formulation. We present our dictionary learning algorithm in Section \ref{sec:algo}. Section \ref{sec:impro} introduces different improvements for this algorithm, and Section \ref{sec:exp} demonstrates experimentally the usefulness of our approach. 

\section{Proposed Approach}
\label{sec:approach}

Given a set of $n$ training image patches of size $m$ pixels, represented by the
columns of a matrix $\bfX=[\bfx_1,\dots,\bfx_n]$ in $\R^{m\times n}$,
the classical dictionary learning formulation, as introduced by~\cite{field} and revisited by \cite{elad,mairal7}, tries to find a dictionary $\D=[\bfd_1,\dots,\bfd_p]$ in $\Real^{m \times p}$ such that each signal $\bfx_i$ can 
be represented by a sparse linear combination of the columns of $\D$.
More precisely, the dictionary $\D$ is learned along with a matrix of
decomposition coefficients $\A = [\alphab_1,\dots,\alphab_n]$ in~$\Real^{p \times n}$ such that $\x_i \approx \D \alphab_i$ for every signal~$\x_i$.
Following \cite{mairal7}, we consider the following formulation:
\begin{equation}
   \min_{\D \in \mathcal{D},\A \in \Real^{p \times n}}
   \frac{1}{n}\sum_{i=1}^n\Big[\frac{1}{2} \|\x_i-\D\alphab_i\|_2^2 + \lambda \|\alphab_i\|_1 \Big],\label{eq:dict_learning}
\end{equation}
where the quadratic term ensures that the vectors $\x_i$ are close to the approximation $\D\alphab_i$, the $\ell_1$-norm induces sparsity in the coefficients $\alphab_i$~(see, e.g., \cite{chen,tibshirani}), and $\lambda$ controls the amount of
regularization.  To prevent the columns of $\D$ from being arbitrarily large
(which would lead to arbitrarily small values of the $\alphab_i$), the
dictionary~$\D$ is constrained to belong to the convex set $\mathcal{D}$ of
matrices in $\Real^{m \times p}$ whose columns have an $\ell_2$-norm less than
or equal to one:
\begin{displaymath}
   \mathcal{D} \defin \{ \D \in \Real^{m \times p} \st \forall j=1,\dots,p,~ \|\bfd_j\|_2 \leq 1 \}.
\end{displaymath}

As will become clear shortly, this constraint is not adapted to dictionaries extracted from epitomes, since overlapping patches cannot be expected to all have the same norm. Thus we introduce an unconstrained formulation equivalent to Eq.~(\ref{eq:dict_learning}):
\begin{equation}
   \min_{\substack{\D \in \R^{m \times n},\\ \A \in \R^{p \times n}}}
   \frac{1}{n}\sum_{i=1}^n\Big[\frac{1}{2} \|\x_i-\D\alphab_i\|_2^2 + \lambda \sum_{j=1}^p \|\bfd_j\|_2 |\alpha_{j,i}| \Big].\label{eq:dict_learning2}
\end{equation}
This formulation removes the constraint $\D \in \mathcal{D}$ from Eq.~(\ref{eq:dict_learning}), and replaces the $\ell_1$-norm by a weighted $\ell_1$-norm. As shown in Appendix~\ref{ap:A}, Eq.~(\ref{eq:dict_learning}) and Eq.~(\ref{eq:dict_learning2}) are equivalent in the sense that a solution of Eq.~(\ref{eq:dict_learning}) is also solution of Eq.~(\ref{eq:dict_learning2}), and for every solution of Eq.~(\ref{eq:dict_learning2}), a solution for Eq.~(\ref{eq:dict_learning}) can be obtained by normalizing its columns to one. To the best of our knowledge, this equivalent formulation is new, and is key to learning an epitome with $\ell_1$-regularization: the use of a convex regularizer (the $\ell_1$-norm) that empirically provides better-behaved dictionaries than $\ell_0$ (where the $\ell_0$ pseudo-norm counts the number of non-zero elements in a vector) for denoising tasks (see Table~\ref{table:denoise}) differentiates us from the ISD formulation of \cite{aharon2}. To prevent degenerate solutions in the dictionary learning formulation with $\ell_1$-norm, it is important to constrain the dictionary elements with the $\ell_2$-norm. Whereas such a constraint can easily be imposed in classical dictionary learning, its extension to epitome learning is not straightforward, and the original ISD formulation is not compatible with convex regularizers. Eq.~(\ref{eq:dict_learning2}) is an equivalent unconstrained formulation, which lends itself well to epitome learning.

We can now formally introduce the general concept of an epitome as a small image of size $\sqrt{M}\times\sqrt{M}$, encoded (for example in row order) as a vector $\bfE$ in $\R^M$. We also introduce a linear operator $\varphi : \Real^M \to \Real^{m \times p}$ that extracts all overlapping patches from the epitome $\bfE$, and rearranges them into the columns of a matrix of $\R^{m\times p}$, the integer $p$ being the number of such overlapping patches. Concretely, we have $p=(\sqrt{M}-\sqrt{m}+1)^2$. In this context, $\varphi(\bfE)$ can be interpreted as a traditional flat dictionary with $p$ elements, except that it is generated by a small number $M$ of parameters compared to the $pm$ parameters of the flat dictionary. 
Our approach thus generalizes to a much wider range of epitomic structures using any mapping $\varphi$ that admits fast projections on $\Im(\varphi)$. The functions $\varphi$ we have used so far are relatively simple, but give a framework that easily extends to families of epitomes, shift-invariant dictionaries, and plain dictionaries. The only assumption we make is that $\varphi$ is a linear operator of rank $M$ (i.e., $\varphi$ is injective). This list is not exhaustive, which naturally opens up new perspectives.
The fact that a dictionary $\D$ is obtained from an epitome is characterized by the fact that $\D$ is in the image $\Im \varphi$ of the linear operator $\varphi$. Given a dictionary $\D$ in $\Im \varphi$, the unique (by injectivity of $\varphi$) epitome representation can be obtained by computing the inverse of $\varphi$ on $\Im \varphi$, for which a closed form using pseudo-inverses exists as shown in Appendix~\ref{ap:B}.

Our goal being to adapt the epitome to the training image patches,
the general minimization problem can therefore be expressed as follows: 
\begin{equation}
   \min_{\substack{\D \in \Im \varphi,\\ \A \in \R^{p \times n}}}
   \frac{1}{n}\sum_{i=1}^n\Big[\frac{1}{2} \|\x_i-\D\alphab_i\|_2^2 + \lambda \sum_{j=1}^p \|\bfd_j\|_2 |\alpha_{j,i}| \Big].\label{eq:epitome_learn}
\end{equation}
There are several motivations for such an approach. As discussed above, the choice of the function $\varphi$ lets us adapt this technique to different problems such as multiple epitomes or any other type of dictionary representation. This formulation is therefore deliberately generic. In practice, we have mainly focused on two simple cases in the experiments of this paper: a single epitome~\cite{jojic} (or image signature dictionary~\cite{aharon2}) and a set of epitomes. Furthermore, we have now come down to a more traditional, and well studied problem: dictionary learning. We will therefore use the techniques and algorithms developed in the dictionary learning literature to solve the epitome learning problem.

\vspace{-.1cm}
\section{Basic Algorithm}
\label{sec:algo}
As for classical dictionary learning, the optimization problem of
Eq.~(\ref{eq:epitome_learn}) is not jointly convex in $(\D,\A)$, but is convex
with respect to $\D$ when $\A$ is fixed and vice-versa.  A block-coordinate
descent scheme that alternates between the optimization of $\D$ and $\A$, while
keeping the other parameter fixed, has emerged as a natural and
simple way for learning dictionaries~\cite{elad,engan}, which
has proven to be relatively efficient when the training set is not too large.
Even though the formulation remains nonconvex and therefore this method is not
guaranteed to find the global optimum, it has proven experimentally to 
be good enough for many tasks~\cite{elad}.

We therefore adopt this optimization scheme as well, and detail the different
steps below. Note that other algorithms such as stochastic gradient descent
(see \cite{aharon2,mairal7}) could be used as well, and in fact can easily be
derived from the material of this section. However, we have chosen not to
investigate these kind of techniques for simplicity reasons. Indeed, stochastic
gradient descent algorithms are potentially more efficient than the
block-coordinate scheme mentioned above, but require the (sometimes non-trivial) tuning
of a learning rate.

\subsection{Step 1: Optimization of $\bfA$ with $\bfD$ Fixed.}
\label{step1}

In this step of the algorithm, $\bfD$ is fixed, so the constraint $\D \in \Im \varphi$ is not involved in the optimization of $\bfA$. Furthermore, note that updating the matrix $\bfA$ consists of solving $n$ independent optimization problems with respect to each column~$\alphab_i$.  For each of them, one has to solve a weighted-$\ell_1$ optimization problem. Let us consider the update of a column~$\alphab_i$ of $\bfA$.

We introduce the matrix $\bfG\defin\diag[\| \bfd_1\|_2,..,\| \bfd_p\|_2]$, and
define $\bfD' = \bfD \bfG^{-1}$. 
If $\bfG$ is non-singular, we show in Appendix~\ref{ap:A} that the relation $\bfa_i^{\prime \star}
= \bfG \bfa_i^\star$ holds, where
\begin{displaymath}
\begin{split}
 \bfa_i^{\prime \star} &= \argmin_{\alphab_i' \in \Real^p}   \frac{1}{2} \| \x_i - \bfD' \alphab_i' \|_F^2 + \lambda \|\alphab_i'\|_1, \ \ \ \text{and}\\
 \bfa_i^{\star} &= \argmin_{\alphab_i \in \Real^p}   \frac{1}{2} \| \x_i - \bfD \alphab_i' \|_F^2 + \lambda \sum_{j=1}^p \|\bfd_j\|_2 |\alpha_{j,i}|. 
\end{split}
\end{displaymath}
This shows that the update of each column can easily be obtained with classical
solvers for $\ell_1$-decomposition problems.  We use to that effect the LARS
algorithm~\cite{efron}, implemented in the software
accompanying~\cite{mairal7}. 

Since our optimization problem is invariant by multiplying $\D$ by a scalar and
$\A$ by its inverse, we then proceed to the following renormalization to ensure
numerical stability and prevent the entries of $\D$ and $\A$ from becoming too
large: we rescale $\bfD$ and $\bfA$ with 
$$s = \min_{j \in \llbracket 1, n
\rrbracket}  \|\bfd_j\|_2, \mbox{ and define } \bfD \leftarrow \frac{1}{s}\bfD
\mbox{ and } \bfA \leftarrow s\bfA.$$
 Since the image of $\varphi$ is a vector
space, $\bfD$ stays in the image of $\varphi$ after the normalization. And as noted before, 
it does not change the value of the objective function.

\subsection{Step 2: Optimization of $\bfD$ with $\bfA$ Fixed.}

We use a projected gradient descent algorithm~\cite{bertsekas} to update $\bfD$.
The objective function $f$ minimized during this step can be written as:
 \begin{equation}
 f(\bfD) \defin \frac{1}{2} \| \bfX-\bfD \bfA \|_F^2 + \lambda \sum_{j=1}^p \|\bfd_j\|_2 \|\bfa^j\|_1, \label{eq:updateD}
\end{equation}
where $\A$ is fixed, and we recall that $\bfa^j$ denotes its $j$-th row. The function
$f$ is differentiable, except when a column of $\D$ is equal to zero,
which we assume without loss of generality not to be the case. 
Suppose indeed that a column $\bfd_j$ of $\D$ is equal to zero. Then, without
changing the value of the cost function of Eq.~(\ref{eq:epitome_learn}), one can
set the corresponding row $\bfa^j$ to zero as well, and it results in 
a function $f$ defined in Eq.~(\ref{eq:updateD}) that does not depend on
$\bfd_j$ anymore. We have, however, not observed such a situation in our experiments.

The function $f$ can therefore be considered as differentiable, and one can
easily compute its gradient as:
\begin{displaymath}
 \nabla f(\D) = - (\bfX-\bfD\bfA)\bfA^T + \D \bfDD,
\end{displaymath}
where $\bfDD$ is defined as $\bfDD \defin
\diag(\lambda\frac{\|\bfa^1\|_1}{\|\bfd_1\|_2},\ldots,\lambda\frac{\|\bfa^p\|_1}{\|\bfd_p\|_2})$.

To use a projected gradient descent, we now need  a method for projecting $\D$
onto the convex set $\Im \varphi$, and the update rule becomes:
\begin{displaymath}
\D \leftarrow  \Pi_{\Im \varphi}[  \D - \rho \nabla f(\D) ],
\end{displaymath}
where $\Pi_{\Im \varphi}$ is the orthogonal projector onto $\Im \varphi$, and $\rho$ is a gradient step, chosen with a line-search rule, such as the Armijo rule~\cite{bertsekas}.

Interestingly, in the case of the single epitome (and in fact in any other
extension where $\varphi$ is a linear operator that extracts some patches from
a parameter vector $\bfE$), this projector admits a closed form: let us
consider the linear operator $\varphi^\ast : \Real^{m \times p} \to \Real^M$,
such that for a matrix $\D$ in $\Real^{m \times p}$, a pixel of the epitome
$\varphi^\ast(\D)$ is the average of the entries of $\D$ corresponding to this
pixel value. We give the formal form of this operator in Appendix~\ref{ap:B}, and show the following results: \\
\hspace*{.5cm} (i) $\varphi^\ast$ is indeed linear, \\
\hspace*{.5cm} (ii) $\Pi_{\Im \varphi}=\varphi \circ \varphi^\ast$. \\
With this closed form of $\Pi_{\Im \varphi}$ in hand, we now have an efficient
algorithmic procedure for performing the projection.
Our method is therefore quite generic, and can adapt to a wide variety of functions $\varphi$. Extending it when $\varphi$ is not linear, but still injective and with an efficient method to project on $\Im \varphi$ will be the topic of future work.

\section{Improvements}
\label{sec:impro}
We present in this section several improvements to our basic framework, which either improve the convergence speed of the algorithm, or generalize the formulation.

\subsection{Accelerated Gradient Method for Updating $\D$.}

A first improvement is to accelerate the convergence of the update of $\D$
using an accelerated gradient technique~\cite{beck,nesterov}.  These methods,
which build upon early works by Nesterov~\cite{nesterov2}, have attracted a lot
of attention recently in machine learning and signal processing, especially
because of their fast convergence rate (which is proven to be optimal among first-order
methods), and their ability to deal with large, possibly nonsmooth problems.

Whereas the value of the objective function with classical gradient descent
algorithms for solving smooth convex problems is guaranteed to decrease with a
convergence rate of $O(1/k)$, where $k$ is the number of iterations, other
algorithmic schemes have been proposed with a convergence rate of $O(1/k^2)$
with the same cost per iteration as classical gradient
algorithms~\cite{beck,nesterov2,nesterov}. 
The difference between these methods and gradient descent algorithms is that
two sequences of parameters are maintained during this iterative procedure, and
that each update uses information from past iterations.
This leads to theoretically better convergence rates, which are often 
also better in practice.

We have chosen here for its simplicity the algorithm FISTA of Beck and Teboulle~\cite{beck}, which includes a practical line-search scheme for
automatically tuning the gradient step.  Interestingly, we have indeed observed
that the algorithm FISTA was significantly faster to converge than the
projected gradient descent algorithm.

\subsection{Multi-Scale Version}

\begin{figure}[h]
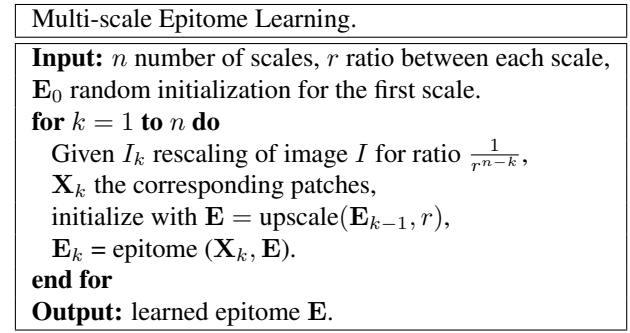

\begin{center}
\begin{tabular}{|l|}
\hline
Multi-scale Epitome Learning.  \\
\hline\hline
\textbf{Input:}  $n$ number of scales, $r$ ratio between each scale, \\
$\bfE_0$ random initialization for the first scale. \\
\textbf{for $k=1$ to $n$ do} \\
\ \ \ Given $I_k$ rescaling of image $I$ for ratio $\frac{1}{r^{n-k}}$, \\
\ \ \ $\bfX_k$ the corresponding patches, \\
\ \ \ initialize with $\bfE=\text{upscale}(\bfE_{k-1},r)$, \\
\ \ \ $\bfE_k$ = epitome ($\bfX_k,\bfE$). \\
\textbf{end for} \\
\textbf{Output:} learned epitome $\bfE$.\\
\hline
\end{tabular}
\end{center}
\caption{Multi-scale epitome learning algorithm.}
\label{table:msl}
\vspace{-.2cm}
\end{figure}

To improve the results without increasing the computing time, we have also
implemented a multi-scale approach that exploits the spatial nature of the
epitome. Instead of directly learning an epitome of size $M$, we first learn an
epitome of a smaller size on a reduced image with corresponding smaller patches,
and after upscaling, we use the resulting epitome as the initialization for the
next scale. We iterate this process in practice two to three times. The procedure is illustrated in Figure \ref{table:msl}.
Intuitively, learning smaller epitomes is an easier task than directly
learning a large one, and such a procedure provides a good initialization for
learning a large epitome.

\subsection{Multi-Epitome Extension}
\label{sec:multiep}

Another improvement is to consider not a single epitome but a family of
epitomes in order to learn dictionaries with some shift invariance, which has
been the focus of recent work~\cite{mailhe,thiagarajan}. Note that
different types of structured dictionaries have also been proposed with the
same motivation for learning shift-invariant features in image classification
tasks~\cite{kavukcuoglu2}, but in a significantly different framework (the
structure in the dictionaries learned in \cite{kavukcuoglu2} comes from a
different sparsity-inducing penalization).

\begin{figure}[hbtp]
\centerline{
\includegraphics[width=\linewidth]{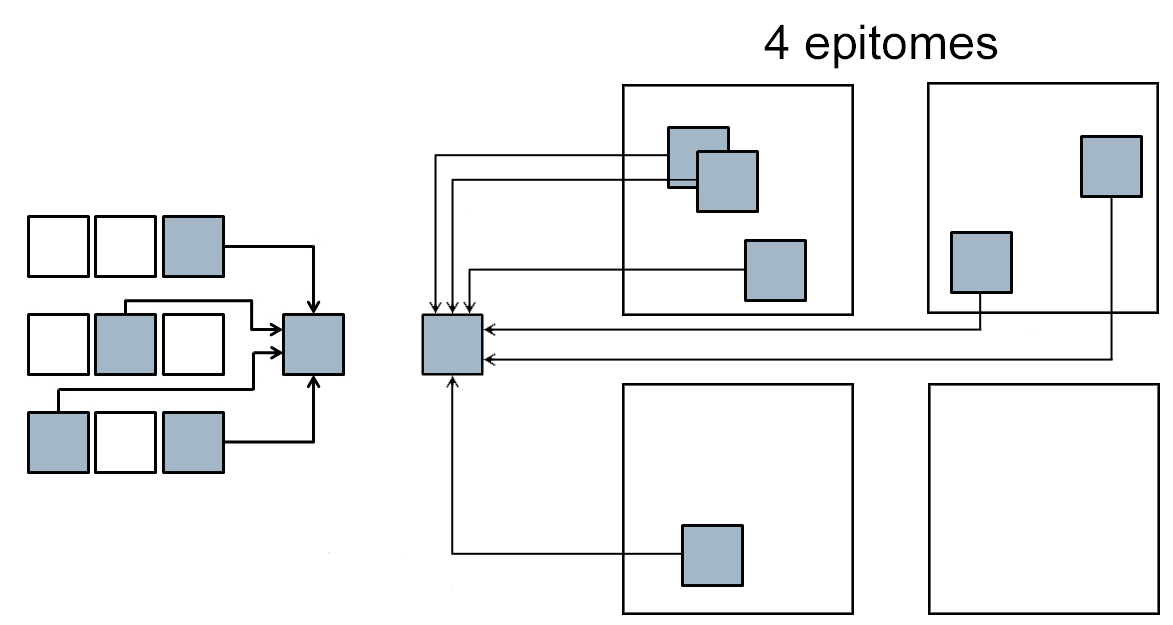}}
\caption{A ``flat'' dictionary (left) vs. a collection of $4$ epitomes (right).
The atoms are extracted from the epitomes and may overlap.}
\label{fig:multiepitomes}
\end{figure}

As mentioned before, we are able to learn a set of $N$ epitomes instead of a
single one by changing the function $\varphi$ introduced earlier.  The vector
$\bfE$ now contains the pixels (parameters) of several small epitomes, and
$\varphi$ is the linear operator that extracts all overlapping patches from all
epitomes.  In the same way, the projector on $\Im \varphi$ is still easy to compute in
closed form, and the rest of the algorithm stays
unchanged.  Other ``epitomic'' structures could easily be used within our
framework, even though we have limited ourselves for simplicity to the case of
single and multiple epitomes of the same size and shape.

The multi-epitome version of our approach can be seen as an interpolation
between classical dictionary and single epitome. Indeed, defining a multitude
of epitomes of the same size as the considered patches is equivalent to working
with a dictionary. Defining a large number a epitomes slightly larger than the
patches is equivalent to shift-invariant dictionaries. 
In Section~\ref{sec:exp}, we experimentally compare these different regimes
for the task of image denoising.

\subsection{Initialization}

Because of the nonconvexity of the optimization problem, the question of the initialization is an important issue in epitome learning.  We have already mentioned a multi-scale strategy to overcome this issue, but for the first scale, the problem remains. Whereas classical flat dictionaries can naturally be initialized with prespecified dictionaries such as overcomplete DCT basis (see~\cite{elad}), the epitome does not admit such a natural choice.
In all the experiences (unless written otherwise), we use as the initialization a single epitome (or a collection of epitomes), common to all experiments, which is learned using our algorithm, initialized with a Gaussian low-pass filtered random image, on a set of $100\,000$ random patches extracted from $5\,000$ natural images (all different from the test images used for denoising).

\section{Experimental Validation}
\label{sec:exp}

\bfg
\includegraphics[width=2.72cm]{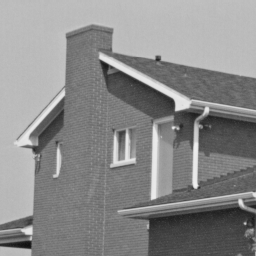}
\includegraphics[width=2.72cm]{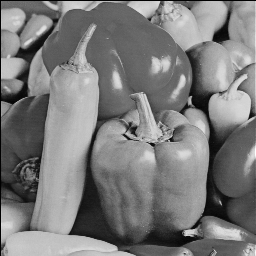}
\includegraphics[width=2.72cm]{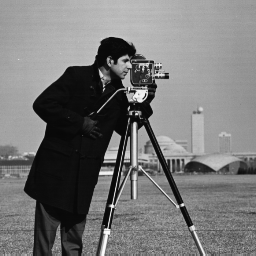}
\includegraphics[width=2.72cm]{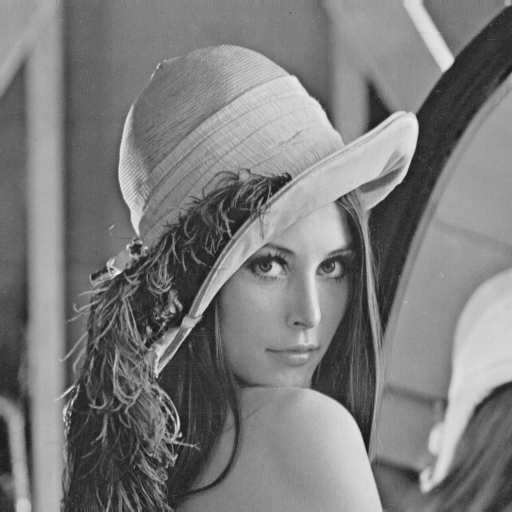}
\includegraphics[width=2.72cm]{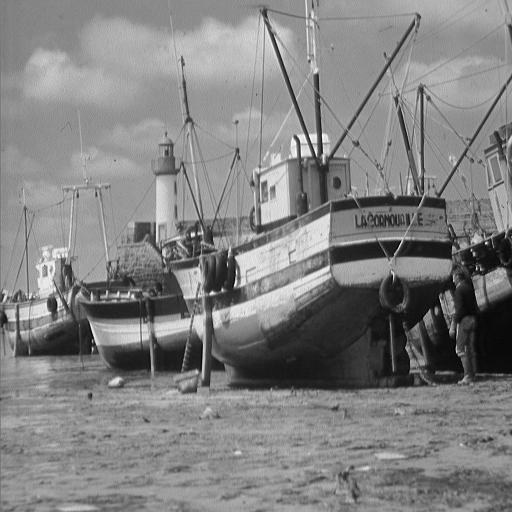}
\includegraphics[width=2.72cm]{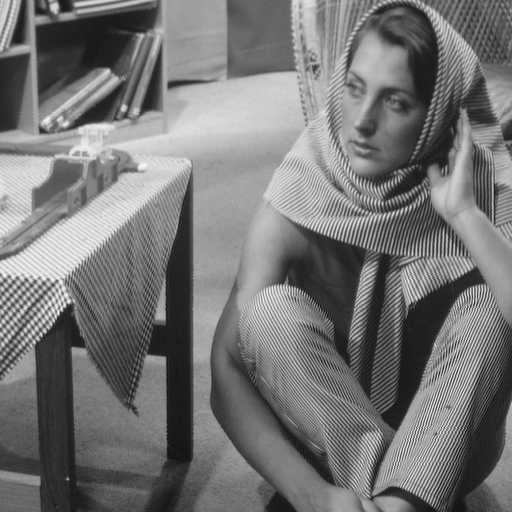}
\caption{House, Peppers, Cameraman, Lena, Boat and Barbara images.}
\label{fig:im}
\vspace{-0.4cm}
\efg

We provide in this section qualitative and quantitative validation. We first study
the influence of the different model hyperparameters on the visual aspect of the 
epitome before moving to an image denoising task.
We choose to represent the epitomes as images in order to visualize more easily the
patches that will be extracted to form the images. Since epitomes contain negative values, they are arbitrarily rescaled between $0$ and $1$ for display.

In this section, we will work with several images, which are shown in Figure~\ref{fig:im}.

\subsection{Influence of the Initialization}

In order to measure the influence of the initialization on the resulting
epitome, we have run the same experience with different initializations.
Figure \ref{fig:init} shows the different results obtained. 

The difference in contrast may be due to the scaling of the data in the displaying process. 
This experiment illustrates that different initializations lead to visually different
epitomes. Whereas this property might not be desirable, the classical dictionary learning
framework also suffers from this issue, but yet has led to successful applications
in image processing~\cite{elad}.

\bfg  
	\includegraphics[width=8.3cm]{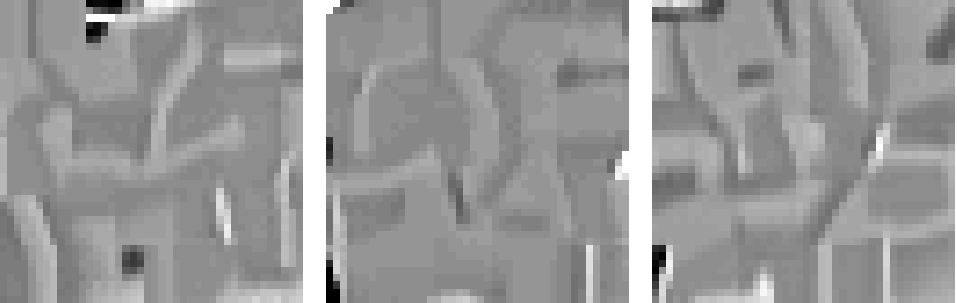}
\caption{Three epitomes obtained on the boat image for different initializations, but all the same parameters. Left: epitome obtained with initialization on a epitome learned on random patches from natural images. Middle and Right: epitomes obtained for two different random initializations.}
\label{fig:init}
\vspace{-0.4cm}
\efg

\vspace{-.3cm}

\subsection{Influence of the Size of the Patches}

The size of the patches seem to play an important role in the visual aspect of the epitome. We illustrate in Figure~\ref{fig:Epi32} an experiment where  pairs of epitome of size $46 \times 46$ are learned with different sizes of patches.

\bfg  
	\includegraphics[width=8.4cm]{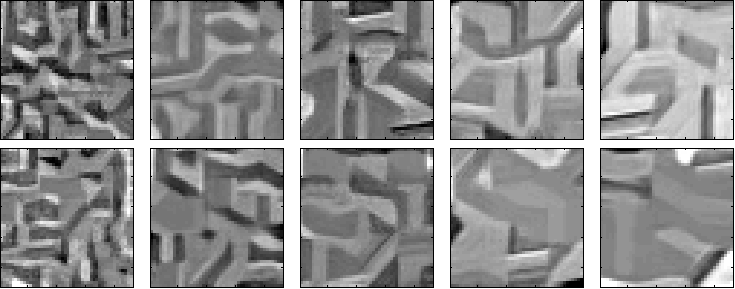}
\caption{Pairs of epitomes of width $46$ obtained for patches of width $6,8,9,10$ and $12$. All other parameters are unchanged. Experiments run with $2$ scales ($20$ iterations for the first scale, $5$ for the second) on the house image. }
\label{fig:Epi32}
\vspace{-0.4cm}
\efg

As we see, learning epitomes with small patches seems to introduce finer details and structures in the epitome, whereas large patches induce epitomes with coarser structures.

\subsection{Influence of the Number of Epitomes}

\begin{figure}
\begin{center}
\includegraphics[width=8cm]{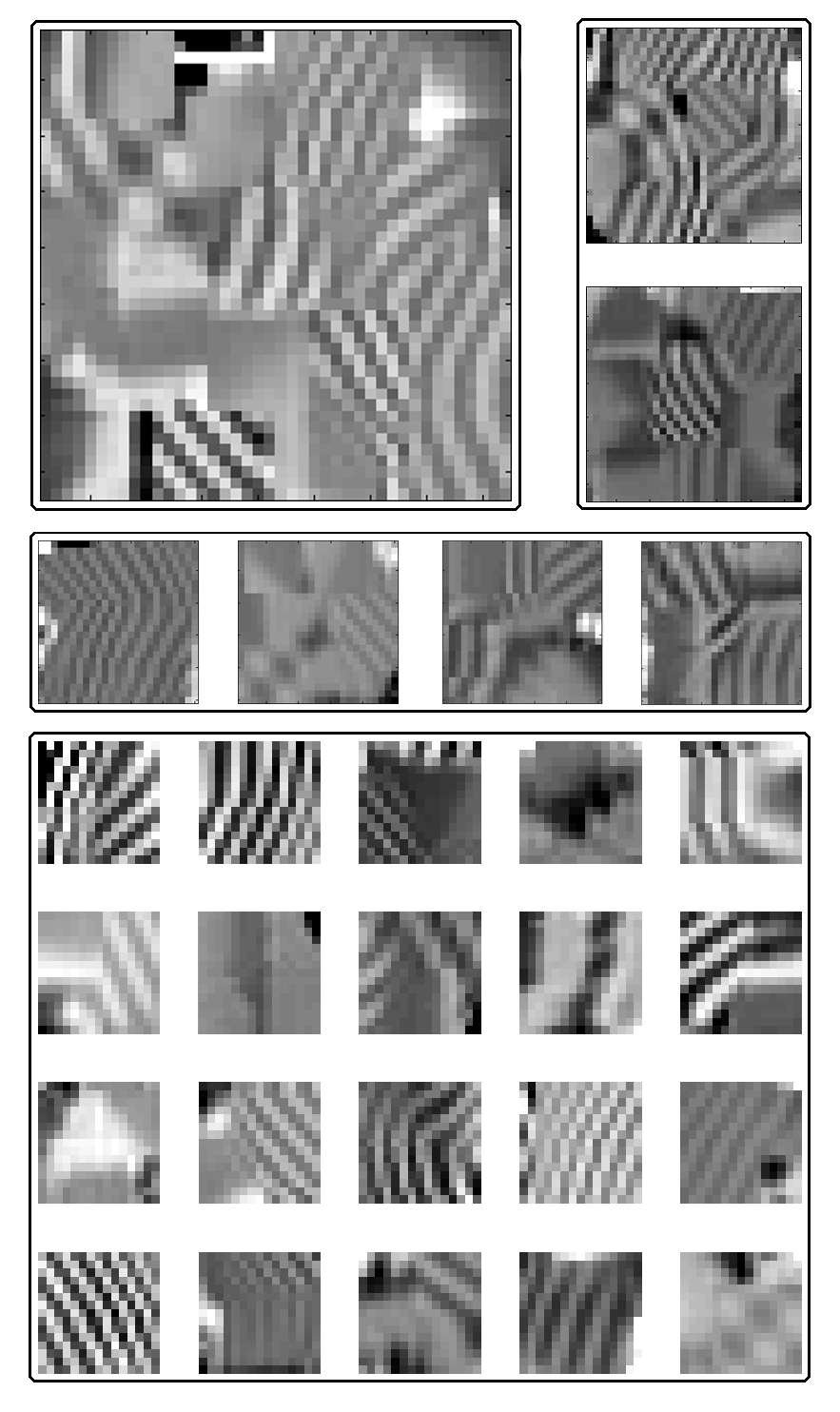}
\caption{$1$, $2$, $4$ and $20$ epitomes learned on the barbara image for the same parameters. They are of sizes $42$, $32$, $25$ and $15$ in order to keep the same number of elements in $\bfD$. They are not represented to scale.}
\label{fig:multiep}
\end{center}
\vspace{-0.7cm}
\end{figure}

We present in this section an experiment where the number of learned epitomes
vary, while keeping the same numbers of columns in $\D$. The $1$, $2$, $4$ and $20$ epitomes learned on the image barbara are shown in Figure~\ref{fig:multiep}. 
When the number of epitomes is small, we observe in the epitomes some discontinuities between texture areas with different visual characteristics, which is not the case when learning several independant epitomes.

\subsection{Application to Denoising}

In order to evaluate the performance of epitome learning in various regimes (single epitome, multiple epitomes), we use the same methodology as~\cite{aharon2} that uses the successful denoising method first introduced by~\cite{elad}. Let us consider first the classical problem of restoring a noisy image~$\y$ in~$\Real^n$ which has been corrupted by a white Gaussian noise of standard deviation~$\sigma$. We denote by $\y_i$ in $\Real^m$ the patch of $\y$ centered at pixel $i$ (with any arbitrary ordering of the image pixels).

The method of~\cite{elad} proceeds as follows:
\vspace{-.15cm}
\begin{itemize}
\item Learn a dictionary $\D$ adapted to all overlapping patches $\y_1,\y_2,\ldots$ from the noisy
image $\y$.
\vspace{-.15cm}
\item Approximate each noisy patch using the learned dictionary with a greedy
algorithm called orthogonal matching pursuit (OMP)~\cite{mallat4} to have a clean
estimate of every patch of $\y_i$ by addressing the following problem
\begin{displaymath}
     \argmin_{\alphab_i \in \R^p} \|\alphab_i\|_0 \st \|\y_i-\D\alphab_i\|_2^2 \leq (C\sigma^2),
\end{displaymath}
where $\D\alphab_i$ is a clean estimate of the patch $\y_i$, $\|\alphab_i\|_0$ is the $\ell_0$ pseudo-norm of $\alphab_i$,
and $C$ is a regularization parameter. Following \cite{elad}, we choose $C=1.15$.
\vspace{-.15cm}
\item Since every pixel in $\y$ admits many clean estimates (one estimate for
every patch the pixel belongs to), average the estimates.
\end{itemize}

\bfg
\includegraphics[width=8.5cm]{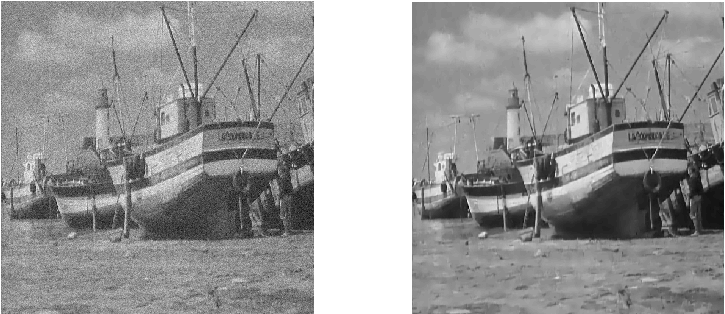}
\caption{Artificially noised boat image (with standard deviation $\sigma=15$), and the result of our denoising algorithm.}
\label{fig:denoising}
\vspace{-0.4cm}
\efg

Quantitative results for single epitome, and multi-scale multi-epitomes are presented in Table~\ref{table:denoise} on six images and five levels of noise. We evaluate the performance of the denoising process by computing the peak signal-to-noise ratio (PSNR) for each pair of images. For each level of noise, we have selected the best regularization parameter $\lambda$ overall the six images, and have then used it all the experiments. The PNSR values are averaged over $5$ experiments with $5$ different noise realizations. The mean standard deviation is of $0.05$dB both for the single epitome and the multi-scale multi-epitomes.

\begin{table}[h]
\begin{center}
\begin{tabular}{| @{\,} c @{\,\,} c @{\,} || @{\,\,} c @{\,\,} | @{\,} c @{\,} | @{\,\,} c @{\,\,} | @{\,} c @{\,} | @{\,\,} c @{\,\,} | @{\,\,} c @{\,\,} |}
\hline
$\sigma$ &   & house & peppers & c.man & barbara & lena & boat \\
\hline 
\multirow{2}{*}{$10$} & I$\bfE$ & 35.98 & \bf{34.52} & \bf{33.90} & \bf{34.41} & \bf{35.51} & \bf{33.70} \\
 & $\bfE$ & 35.86  & 34.41 & 33.83 & 34.01 & 35.43 & 33.63 \\
\hline
\multirow{2}{*}{$15$} & I$\bfE$ & \bf{34.45} & \bf{32.50} & \bf{31.65} & \bf{32.23} & \bf{33.74} & \bf{31.81}\\
 & $\bfE$ & 34.32 & 32.36& 31.59 & 31.84 & 33.66 & 31.75 \\
\hline
\multirow{2}{*}{$20$} & I$\bfE$ & \bf{33.18} & \bf{31.00} & \bf{30.19} & \bf{30.69} & \bf{32.42} & \bf{30.45}\\
 & $\bfE$ & 33.08 & 30.93 & 30.11 & 30.33 & 32.35 & 30.37 \\
\hline
\multirow{2}{*}{$25$} & I$\bfE$ & \bf{32.02} & \bf{29.82} & \bf{29.08} & \bf{29.49} & \bf{31.36} & \bf{29.36}\\
 & $\bfE$ & 31.96 & 29.77 & 29.01 & 29.14 & 31.29 & 29.30 \\
\hline
\multirow{2}{*}{$50$} & I$\bfE$ & 27.83 & 26.06 & 25.57 & \bf{25.04} & \bf{27.90} & 26.01 \\
 & $\bfE$ & \bf{27.83} & \bf{26.07} & \bf{25.60} & 24.86 & 27.82 & \bf{26.02} \\
\hline
\end{tabular}
\end{center}
\caption{PSNR Results. 
First Row: $20$ epitomes of size $ 7 \times 7$ learned with 3 scales (I$\bfE$: improved epitome); 
Second row: single epitome of size $42 \times 42$ ($\bfE$). Best results are in bold.} 
\label{table:denoise}
\end{table}

\begin{table}[h]
\begin{center}
\begin{tabular}{|@{\,}c@{\,}||@{\,\,}c@{\,\,}|@{\,\,}c@{\,\,}|@{\,\,}c@{\,\,}|@{\,\,}c@{\,\,}||@{\,\,}c@{\,\,}|@{\,\,}c@{\,\,}|@{\,\,}c@{\,\,}|}
\hline
$\sigma$ & I$\bfE$ & $\bfE$ & \cite{aharon2} & \cite{jojic} & \cite{elad} & \cite{dabov2} & \cite{mairal8} \\
\hline 
$10$ & $34.83$ & $34.67$ & $34.71$ & $28.83$  & $34.76$ & $35.24$ & $35.32$ \\
$15$ & $32.95$ & $32.79$ & $32.84$ & $28.92$ & $32.87$ & $33.43$ & $33.50$ \\
$20$ & $31.55$ & $31.41$ & $31.36$ & $28.55$ & $31.52$ & $32.15$ & $32.18$ \\
$25$ & $30.41$ & $30.29$ & $29.99$ & $28.12$ & $30.42$ & $31.15$ & $31.11$ \\
$50$ & $26.57$ & $26.52$ & $25.91$ & $25.21$ & $26.66$ & $27.69$ & $27.87$ \\
\hline
mean & $31.26$ & $31.14$ & $30.96$ & $27.93$ & $31.25$ & $31.93$ & $32.00$ \\
\hline
\end{tabular}
\end{center}
\caption{Quantitative comparative evaluation. PSNR values are averaged over $5$ images. We compare ourselves to two previous epitome learning based algorithms: ISD (\cite{aharon2}) and epitomes by Jojic, Frey and Kannan (\cite{jojic} as reported in \cite{aharon2}), and to three more elaborate dictionary learning based algorithms K-SVD (\cite{elad}), BM3D (\cite{dabov2}), and LSSC (\cite{mairal8}).}
\vspace{-0.2cm}
\end{table}

We see from this experiment that the formulation we propose is competitive compared to the one of \cite{aharon2}. Learning multi epitomes instead of a single one seems to provide better results, which might be explained by the lack of flexibility of the single epitome representation.
Evidently, these results are not as good as recent state-of-the-art denoising algorithms such as \cite{dabov2,mairal8} which exploit more sophisticated image models. But our goal is to illustrate the performance of epitome learning on an image reconstruction task, in order to better understand these formulations.

\section{Conclusion}
\label{sec:ccl}

We have introduced in this paper a new formulation and an efficient algorithm
for learning epitomes in the context of sparse coding, extending the work of
Aharon and Elad~\cite{aharon2}, and unifying it with recent work on
shift-invariant dictionary learning. Our approach is generic, can interpolate 
between these two regimes, and can possibly be applied to other formulations.
Future work will extend our framework to
the video setting, to other image processing tasks such as inpainting, and to
learning image features for classification or recognition tasks, where shift
invariance has proven to be a key property to achieving good
results~\cite{kavukcuoglu2}.  Another direction we are pursuing is to find a
way to encode other invariant properties through different mapping functions
$\varphi$.


\paragraph{Acknowledgments.} \hspace{-.3cm} This work was partly supported \hspace{-.1cm} by the European \hspace{-.1cm} Community under the ERC \hspace{-.1cm} grants "VideoWorld" and "Sierra".

\appendix
\section{Appendix: $\ell_1$-Norm and Weighted $\ell_1$-Norm}
\label{ap:A}

In this appendix, we will show the equivalence between the two minimization problems introduced in section \ref{step1}.

Let us denote \beq F(\bfD,\bfa) = \frac{1}{2} \| \bfx-\bfD \bfa \|_2^2  + \frac{\lambda}{2}  \sum_{j=1}^p \|\bfd_j\|_2 |\alpha_{j}|, \eeq \beq \text{and} ~~\ G(\bfD,\bfa)=  \frac{1}{2} \left| \left| \bfx - \bfD \bfa \right| \right|_2^2 + \lambda \|\bfa\|_1 .\eeq

Let us define $\bfa' \in \R^{p}$ and $\bfD' \in \R^{m\times p}$ such that $\bfD' = \bfD \bfG^{-1}$, and $\bfa' = \bfG \bfa$, where $\bfG=\diag[\| \bfd_1\|_2,..,\| \bfd_p\|_2]$. The goal is to show that $\bfa'^{\star}=\bfG\bfa^{\star},$ where:
$$\bfa^{\star} =\argmin_{\bfa} F(\bfD,\bfa),  \ \ \text{and} \ \ \bfa'^{\star} = \argmin_{\bfa'} G(\bfD',\bfa').$$

We clearly have: $\bfD\bfa=\bfD'\bfa'$.
Furthermore, since $\bfG \bfa=\bfa'$, we have: $ \forall j = 1,\ldots,p, ~~\ \|\bfd_j\|_2  |\alpha_j| = |\alpha'_j|.$

Therefore, \beq F(\bfD,\bfa) =G(\bfD',\bfa').\eeq

Moreover, since for all $\D$, $\D'$ is in the set ${\mathcal D}$, we have shown the equivalence between Eq.~(\ref{eq:dict_learning}) and Eq.~(\ref{eq:dict_learning2}).

\section{Appendix: Projection on $\Im \varphi$}
\label{ap:B}

In this appendix, we will show how to compute the orthogonal projection on the vector space $\Im \varphi$. 
Let us denote by $\bfR_i$ the binary matrix in $\{0,1\}^{m \times M}$ that extracts the $i$-th patch from $\bfE$.
Note that with this notation, the matrix $\bfR_i$ is a binary $M \times m$ matrix corresponding to a linear operator that takes a patch of size $m$ and place it at the location $i$ in an epitome of size $M$ which is zero everywhere else.
We therefore have $\varphi(\bfE) = [\bfR_1\bfE,\ldots,\bfR_p\bfE]$.

We denote by $\varphi^\ast : \Real^{m \times p} \to \Real^M$ the linear operator defined as
\begin{displaymath}
\varphi^\ast(\D) = (\sum_{j=1}^p\bfR_j^T\bfR_j)^{-1}(\sum_{j=1}^p \bfR_i^T\D),
\end{displaymath}
which creates an epitome of size $M$ such that each pixel contains the average of the corresponding entries in $\D$. Indeed, the $M \times M$ matrix $(\sum_{j=1}^p\bfR_j^T\bfR_j)^{-1}$ is diagonal and the entry $i$ on the diagonal is the number of entries in $\D$ corresponding to the pixel $i$ in the epitome. 

Denoting by 
$$\bfR \defin \left[
\begin{array}{c}
\bfR_1 \\
\vdots \\
\bfR_p \\
\end{array} \right],
$$
which is a $mp \times M$ matrix, we have
$\text{vec}(\varphi(\bfE)) = \bfR\bfE$, where $\text{vec}(\D)\defin [\bfd_1^T,\ldots,\bfd_p^T]^T$, which is the vector of size $mp$ obtained by concatenating the columns of $\D$,
and also
$\varphi^\ast(\D) = (\bfR^T\bfR)^{-1}\bfR^T\text{vec}(\D)$.

Since $\text{vec}(\Im \varphi) = \Im \bfR$ and $\text{vec}(\varphi(\varphi^\ast(\D)))= \bfR(\bfR^T\bfR)^{-1}\bfR^T\text{vec}(\D)$, which is an orthogonal projection onto $\Im \bfR$,
it results the two following properties which are useful in our framework and classical in signal processing with overcomplete representations (\cite{mallat}):
\begin{itemize}
\item $\varphi^\ast$ is the inverse function of $\varphi$ on $\Im \varphi$: $\varphi^\ast \circ \varphi = \Id$.
\item $(\varphi \circ \varphi^\ast)$ is the orthogonal projector on $\Im \varphi$.
\end{itemize}

{\bibliographystyle{ieee}
\bibliography{biblio}

\begin{thebibliography}{10}\itemsep=-1pt

\bibitem{aharon2}
M.~Aharon and M.~Elad.
\newblock Sparse and redundant modeling of image content using an
  image-signature-dictionary.
\newblock {\em SIAM Journal on Imaging Sciences}, 1(3):228--247, July 2008.

\bibitem{beck}
A.~Beck and M.~Teboulle.
\newblock {A fast iterative shrinkage-thresholding algorithm for linear inverse
  problems}.
\newblock {\em SIAM Journal on Imaging Sciences}, 2(1):183--202, 2009.

\bibitem{bertsekas}
D.~P. Bertsekas.
\newblock {\em Nonlinear programming}.
\newblock Athena Scientific Belmont, 1999.

\bibitem{chen}
S.~S. Chen, D.~L. Donoho, and M.~A. Saunders.
\newblock Atomic decomposition by basis pursuit.
\newblock {\em SIAM Journal on Scientific Computing}, 20(1):33--61, 1998.

\bibitem{cheung}
V.~Cheung, B.~Frey, and N.~Jojic.
\newblock Video epitomes.
\newblock In {\em Proc. CVPR}, 2005.

\bibitem{chu}
X.~Chu, S.~Yan, L.~Li, K.~L. Chan, and T.~S. Huang.
\newblock Spatialized epitome and its applications.
\newblock In {\em Proc. CVPR}, 2010.

\bibitem{dabov2}
K.~Dabov, A.~Foi, V.~Katkovnik, and K.~Egiazarian.
\newblock {Image Denoising by Sparse 3-D Transform-Domain Collaborative
  Filtering}.
\newblock {\em IEEE Transactions on Image Processing}, 16(8):2080--2095, 2007.

\bibitem{efron}
B.~Efron, T.~Hastie, I.~Johnstone, and R.~Tibshirani.
\newblock Least angle regression.
\newblock {\em Annals of statistics}, 32(2):407--499, 2004.

\bibitem{elad}
M.~Elad and M.~Aharon.
\newblock Image denoising via sparse and redundant representations over learned
  dictionaries.
\newblock {\em IEEE Transactions on Image Processing}, 54(12):3736--3745,
  December 2006.

\bibitem{engan}
K.~Engan, S.~O. Aase, and J.~H. Husoy.
\newblock Frame based signal compression using method of optimal directions
  ({MOD}).
\newblock In {\em IEEE International Conference on Acoustics, Speech and Signal
  Processing (ICASSP)}, 1999.

\bibitem{jojic}
N.~Jojic, B.~Frey, and A.~Kannan.
\newblock Epitomic analysis of appearance and shape.
\newblock In {\em Proc. ICCV}, 2003.

\bibitem{kavukcuoglu2}
K.~Kavukcuoglu, M.~Ranzato, R.~Fergus, and Y.~LeCun.
\newblock Learning invariant features through topographic filter maps.
\newblock In {\em Proc. CVPR}, 2009.

\bibitem{mailhe}
B.~Mailh{\'e}, S.~Lesage, R.~Gribonval, F.~Bimbot, and P.~Vandergheynst.
\newblock {Shift-invariant dictionary learning for sparse representations:
  extending K-SVD}.
\newblock In {\em Proc. EUSIPCO}, 2008.

\bibitem{mairal7}
J.~Mairal, F.~Bach, J.~Ponce, and G.~Sapiro.
\newblock Online learning for matrix factorization and sparse coding.
\newblock {\em Journal of Machine Learning Research}, 11:19--60, 2010.

\bibitem{mairal8}
J.~Mairal, F.~Bach, J.~Ponce, G.~Sapiro, and A.~Zisserman.
\newblock Non-local sparse models for image restoration.
\newblock In {\em Proc. ICCV}, 2009.

\bibitem{mallat}
S.~Mallat.
\newblock {\em A Wavelet Tour of Signal Processing, Second Edition}.
\newblock {Academic Press, New York}, 1999.

\bibitem{mallat4}
S.~Mallat and Z.~Zhang.
\newblock Matching pursuit in a time-frequency dictionary.
\newblock {\em IEEE Transactions on Signal Processing}, 41(12):3397--3415,
  1993.

\bibitem{nesterov2}
Y.~Nesterov.
\newblock {A method for solving the convex programming problem with convergence
  rate $O(1/k^2)$}.
\newblock {\em Dokl. Akad. Nauk SSSR}, 269:543--547, 1983.

\bibitem{nesterov}
Y.~Nesterov.
\newblock Gradient methods for minimizing composite objective function.
\newblock Technical report, Center for Operations Research and Econometrics
  (CORE), Catholic University of Louvain, 2007.

\bibitem{ni}
K.~Ni, A.~Kannan, A.~Criminisi, and J.~Winn.
\newblock Epitomic location recognition.
\newblock In {\em Proc. CVPR}, 2008.

\bibitem{field}
B.~A. Olshausen and D.~J. Field.
\newblock Sparse coding with an overcomplete basis set: A strategy employed by
  {V}1?
\newblock {\em Vision Research}, 37:3311--3325, 1997.

\bibitem{peyre}
G.~Peyr\'e.
\newblock Sparse modeling of textures.
\newblock {\em Journal of Mathematical Imaging and Vision}, 34(1):17--31, 2009.

\bibitem{thiagarajan}
J.~Thiagarajan, K.~Ramamurthy, and A.~Spanias.
\newblock Shift-invariant sparse representation of images using learned
  dictionaries.
\newblock In {\em IEEE International Workshop on Machine Learning for Signal
  Processing}, 2008.

\bibitem{tibshirani}
R.~Tibshirani.
\newblock Regression shrinkage and selection via the {L}asso.
\newblock {\em Journal of the Royal Statistical Society. Series B},
  58(1):267--288, 1996.

\end{thebibliography}
}

\end{document}